\pdfoutput=1

\documentclass[11pt]{article}

\usepackage[preprint]{acl}

\usepackage{times}
\usepackage{latexsym}
\usepackage{fontawesome}
\usepackage{bbding}

\usepackage[T1]{fontenc}
\usepackage{amssymb}
\usepackage{pifont}

\usepackage[utf8]{inputenc}

\usepackage{microtype}

\usepackage{inconsolata}

\usepackage{tcolorbox}
\usepackage{graphicx}
\usepackage{amsmath}
\usepackage{enumitem}
\usepackage{booktabs}
\usepackage{multicol,multirow}
\usepackage{adjustbox}
\usepackage{makecell}
\usepackage{longtable}

\usepackage{color, colortbl}
\usepackage{xcolor}
\definecolor{LightCyan}{rgb}{0.8, 0.9, 1}

\tcbuselibrary{skins}
\newcommand{\ourmethod}{\textsc{PerRecBench}}

\title{Can Large Language Models Understand Preferences in \\Personalized Recommendation?}

\author{Zhaoxuan Tan$^{\clubsuit}$, Zinan Zeng$^{\diamondsuit}$, Qingkai Zeng$^{\clubsuit}$, Zhenyu Wu$^{\clubsuit\diamondsuit}$, \\
\textbf{Zheyuan Liu}$^{\clubsuit}$, \textbf{Fengran Mo}$^{\spadesuit}$, \textbf{Meng Jiang}$^{\clubsuit}$ \\
$^{\clubsuit}$University of Notre Dame, $^{\diamondsuit}$Xi'an Jiaotong University, $^{\spadesuit}$Université de Montréal \\
\texttt{\{ztan3, mjiang2\}@nd.edu}
}

\newtcolorbox{prompt}[1]{
    enhanced,
    drop shadow=black!5!white,
    left=4mm,
    right=4mm,
    top=2mm,
    bottom=2mm,
    boxsep=0mm,
    rounded corners,
    title=#1,    fontupper=\footnotesize\linespread{0.9}\fontfamily{lmr}\selectfont,
    }

\begin{document}
\maketitle
\begin{abstract}

Large Language Models (LLMs) excel in various tasks, including personalized recommendations. Existing evaluation methods often focus on rating prediction, relying on regression errors between actual and predicted ratings. However, user rating bias and item quality, two influential factors behind rating scores, can obscure personal preferences in user-item pair data. To address this, we introduce \ourmethod{}, disassociating the evaluation from these two factors and assessing recommendation techniques on capturing the personal preferences in a grouped ranking manner. We find that the LLM-based recommendation techniques that are generally good at rating prediction fail to identify users' favored and disfavored items when the user rating bias and item quality are eliminated by grouping users. With \ourmethod{} and 19 LLMs, we find that while larger models generally outperform smaller ones, they still struggle with personalized recommendation. Our findings reveal the superiority of pairwise and listwise ranking approaches over pointwise ranking, \ourmethod{}’s low correlation with traditional regression metrics, the importance of user profiles, and the role of pretraining data distributions. We further explore three supervised fine-tuning strategies, finding that merging weights from single-format training is promising but improving LLMs’ understanding of user preferences remains an open research problem.\footnote{Code and data are available at \url{https://github.com/TamSiuhin/PerRecBench}.}


\end{abstract}

\section{Introduction}
Personalization tailors system interactions, content, or recommendations to individual users by analyzing their behavior, preferences, and characteristics \cite{tan2023user, zhang2024personalization}. It is critical in domains such as content recommendation \cite{qian2013personalized, baek2023knowledge}, user simulation \cite{dejescu2023approaches}, personalized chatbots \cite{srivastava2020personalized}, user profiling \cite{gu2020hierarchical, gao2023chat}, healthcare \cite{goldenberg2021personalization}, and education \cite{pratama2023revolutionizing}. 
Large Language Models (LLMs) excel in diverse natural language tasks, showcasing emergent abilities \cite{wei2022emergent, lu2023emergent}. To align LLM outputs with individual user preferences, personalization has become a key research focus, necessitating benchmarks for evaluation \cite{li2024personalized, sun2023take, sun2024beyond}. Personalized recommendation, with its abundant user behavior data and preference signals, is widely adopted as a proxy for assessing LLM personalization \cite{kang2023llms}.
 \begin{figure}[t]
    \centering
    \includegraphics[width=1\linewidth]{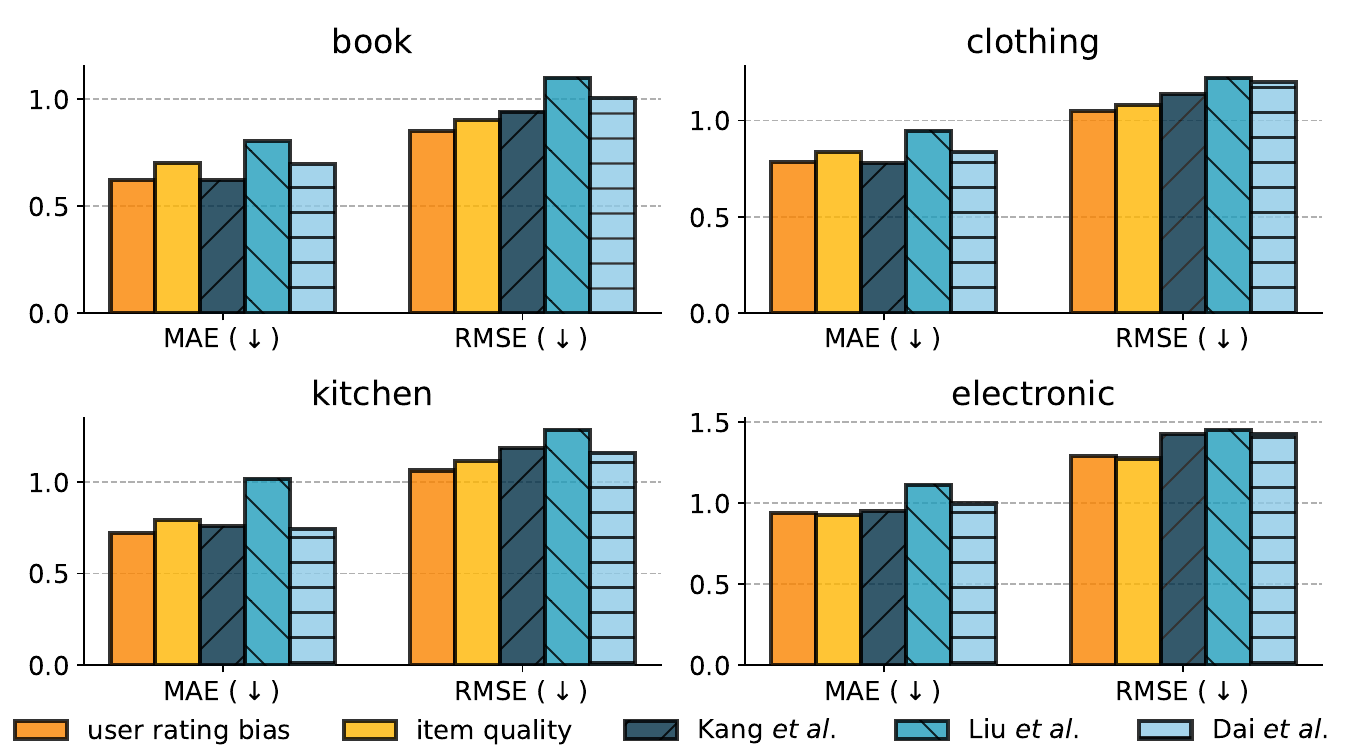}
    \caption{MAE and RMSE performance of user rating bias (average user rating history), item quality (average item rating), and existing LLM-based personalization methods. Simple averages of user rating history and item quality, which do not consider individual preferences, achieve state-of-the-art performance across four shopping domains, questioning the validity of MAE and RMSE for evaluating \textit{personalization}.}
    \label{fig:teaser}
\end{figure}

Personalized recommendation evaluation can be typically categorized into rating-based and ranking-based paradigms. In rating-based evaluation, models predict a user’s rating for an item and calculate regression errors such as MAE and RMSE against actual ratings. However, user rating bias and query item quality are two influential factors behind the rating scores from a user to an item, which might prevent personal preferences in user-item pairs data from being learned. We hypothesize that naive methods that average user rating history (user rating bias) or averaged item rating (item quality) can achieve competitive MAE and RMSE scores. To validate this, we sampled 1,000 user behaviors from the Amazon review dataset \cite{hou2024bridging} across books, clothing, kitchen, and electronics domains and compared the performance with existing LLM-based personalized recommendation methods \cite{kang2023llms, liu2023chatgpt, dai2023uncovering}. Results in Figure \ref{fig:teaser} show that relying solely on statistics like user bias and item quality can achieve strong regression results without incorporating personalized preferences, a pervasive issue in recommendation evaluation. The top-k recommendation involves predicting a user's top-k favorite items based on the user's history, considering a recommendation successful if the predicted item is reviewed and rated highly. However, this approach relies on incomplete signals, as it samples distractors from unreviewed items. These distractors are not inherently poor recommendations, as their exposure to users remains unknown. Effective personalization evaluation should focus on observed signals, distinguishing between low-rated and high-rated preference signals from the user.




To isolate personalization in recommendation evaluation from user rating bias and item quality, we introduce \ourmethod{}, a benchmark that assesses personalization based on observed user preferences in a grouped ranking framework. Specifically, models rank users within a group by their preferences for a shared query item.
To eliminate user rating bias, we define relative rating as the actual rating minus the user’s average rating, where a positive relative rating indicates the user prefers the item over other purchased items. To control for item quality, we group users who purchased the same item within a short timeframe, ensuring consistent item quality within each group. Ground truth rankings are derived by ordering users in each group based on their relative ratings for the shared item.
\ourmethod{} evaluates model performance using pointwise, pairwise, and listwise ranking methods to rank users and measure correlations with ground truth rankings. While input is identical across users in a group, outputs are expected to reflect personalized preferences based on individual profiles and histories. By focusing on observed signals and controlling variables on user rating bias and item quality, \ourmethod{} ensures reliable assessment of personalization.

Using data from Amazon review \cite{hou2024bridging}, we constructed \ourmethod{} with 600 user groups, including 200 groups each with 2, 3, and 4 users to represent increasing levels of difficulty. Benchmarking 19 off-the-shelf LLMs revealed generally unsatisfactory performance, with open-source models exceeding 100B parameters approaching the performance of proprietary models. Among these, \textsc{Claude-3.5-Sonnet} performs best overall. While larger LLMs generally outperformed smaller ones, scaling laws did not consistently hold, as increased model size did not always translate to better performance. Moreover, the low correlation between \ourmethod{} results and MAE/RMSE confirms that personalization is distinct from traditional rating regression tasks. Further analysis highlights the importance of textual user profiles, domain relevance, and shot/retrieval $k$ settings on model performance.

We also investigate three supervised fine-tuning (SFT) strategies to enhance personalization: \textit{single-format training}, \textit{joint training}, and \textit{weight merging}. Single-format training improves task performance and cross-task generalization, while weight merging achieves the best results on \ourmethod{}. However, developing LLMs with robust personalization capabilities remains an open challenge.


In summary, our contributions include introducing \ourmethod{}, the first recommendation benchmark specifically designed to evaluate personalization by removing user rating bias and item quality through observed preference signals, and exploring initial strategies to tackle challenges in LLM-based personalized recommendations.


\begin{figure*}
    \centering
    \includegraphics[width=0.9\linewidth]{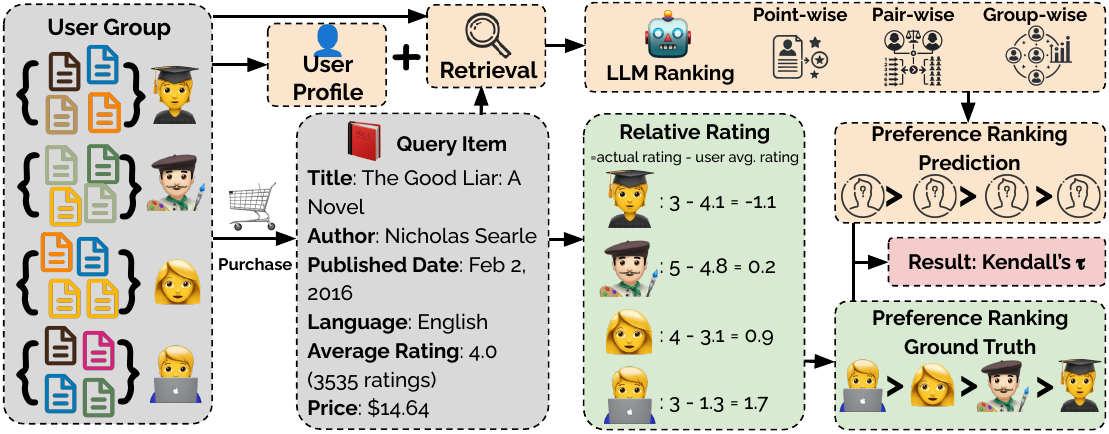}
    \caption{Overview of \ourmethod{}, where the LLM ranks user preferences for a query item using pointwise, pairwise, and listwise prompting. The ground-truth ranking is derived from relative ratings, calculated as the user’s actual rating minus their average rating, to mitigate user rating bias. Finally, Kendall’s tau is computed between the predicted ranking and the ground-truth ranking to evaluate performance.}
    \label{fig:overview}
\end{figure*}

\section{\ourmethod{}}
To assess whether LLMs can capture users’ personalized preferences rather than relying on rating bias or item quality, we introduce the \ourmethod{} Benchmark (Figure \ref{fig:overview}). We first select user groups from diverse shopping domains with varying sizes (\S \ref{sec:user_group}). Next, we evaluate personalization by ranking users by their preferences towards query item using LLM-based ranking methods, including pointwise, pairwise, and listwise approaches (\S \ref{sec:ranking}). Finally, we define evaluation metrics tailored to \ourmethod{} (\S \ref{sec:metric}).

\subsection{User Group Selection}
\label{sec:user_group}
Let $U$ be the set of all users, and let $\mathcal{H}_u = \{(x_u^t, y_u^t)\}$ 
denote the historical behavior of user $u\in \mathcal{U}$, where $x_u^t$ is the item purchased at timestamp $t$, and $y_u^t$ is the corresponding rating.
The goal of user group selection is to select a query item $q$ and the corresponding user subset $\mathcal{U}^*\subseteq \mathcal{U}$ that meet the following criteria:

\paragraph{Temporal Item Co-Purchase.}
All users in the group must have purchased the query item $q$ within a specific time interval $[t_0, t_0 + \Delta t]$, ensuring consistent item quality. Thus, users share the same query item $q$ but have distinct histories. Formally:
\begin{align*}
\forall u \in \mathcal{U}^*: \exists t \in [t_0, t_0 + \Delta t], \; x_u^t = q.
\end{align*}

\paragraph{Active Users.}

Each user must have a sufficient rating history to enable effective personalization. A user is considered active if their history prior to purchasing $q$ exceeds a threshold $\gamma$:
\begin{align*}
     \forall u \in \mathcal{U}^*: \big|\{(x_u^{t'}, y_u^{t'}) \in \mathcal{H}_u, t' < t(q_u)\}\big| > \gamma,
\end{align*}
where $t(q_u)$ denotes the timestamp that user $u$ purchased query item $q$, we set the active level threshold $\gamma$ to 20 by default. 

\paragraph{Significant Relative Rating Difference.}
To capture meaningful preference signals, users within the group must demonstrate distinct preferences for the query item $q$. We address user rating bias by introducing relative ratings. Let $y(q_u)$ denote user $u$'s rating for query item $q$, and define the relative rating as $\tilde{y}(q_u) = y(q_u) - \bar{y}_u$, where $\bar{y}_u$ represents user $u$'s average rating, eliminating the user rating bias. A positive relative rating indicates that the user prefers the query item compared to their average rating, while a negative value suggests the opposite. To ensure distinguishable preferences within the user group, the relative rating difference between any two users must exceed a threshold $\lambda$, set to 0.6 by default:
\begin{align*}
    \forall u, v \in \mathcal{U}^*, u \neq v: \; |\tilde{y}(q_u) - \tilde{y}(q_v)| > \lambda.
\end{align*}

Using these criteria, we constructed \ourmethod{}, comprising 663 user groups with five users per group across book, clothing, kitchen, and electronic domains. To create a progressive testbed, we randomly down-sampled 200 user groups with sizes of 2, 3, and 4, representing ascending levels of difficulty.

\subsection{LLM-based Ranking Methods}
\label{sec:ranking}
\ourmethod{} evaluates LLMs’ ability to rank users’ preferences for a shared query item. Formally, the task requires the LLM to predict a ranking $r$ for $u \in \mathcal{U}^*$ and compare it to the ground truth ranking $r^*$ based on user preferences. To comprehensively assess LLMs’ personalization capabilities, we adopt multiple ranking approaches, including pointwise rating prediction, pairwise ranking, and listwise ranking, to evaluate their effectiveness in modeling user preferences.


\paragraph{Pointwise Rating Prediction}
Given a single user $u$ with rating history $\mathcal{H}_u$, we feed the top-$k$ most relevant user's behavior history \textit{w.r.t.} query item $q$ and user profile $p_u$ to the LLM, asking it to predict user's rating $s_u$ for the query item $q$.
\begin{align*}
    s_u = \mathrm{LLM}(\phi_{pt}(q, \mathcal{D}_u^q, p_u)),
\end{align*}
where $\phi_{pt}$ is the pointwise rating prediction prompt template, and $p_u=\mathrm{LLM}(\mathcal{H}_u)$ denotes the textual user profile generated by an instruction-tuned LLM. The retrieved user history $\mathcal{D}_u^q=\mathcal{R}(q_u, \mathcal{H}_u^{<t(q_u)}, k)$ represents the top-$k$ relevant user history prior to the timestamp of $q_u$, with $\mathcal{R}$ as the retriever. Using the predicted ratings for all users in the group, we compute each user's predicted relative rating $\tilde{s}_u = s_u-\bar{y}_u$, where $\bar{y}_u$ is the user's average rating. Users are then ranked based on their predicted relative rating $r = \mathrm{argsort}(\{\tilde{s}_u, u\in \mathcal{U}^*\})$.

\paragraph{Pairwise Ranking}
As shown in prior research \cite{qin-etal-2024-large, sun-etal-2023-chatgpt}, LLMs can effectively perform text ranking through pairwise comparisons. Similarly, we use pairwise ranking paradigm to rank users based on their preferences on the query item. In pairwise ranking, the fundamental unit is the comparison of user preferences for the same query item. The pairwise comparison function $f$ between user $u_i$ and $u_j$ is:
\begin{align*}
    f(u_i, u_j) = \mathrm{LLM}[\phi_{pr}(q, (\mathcal{D}_{u_i}^q, p_{u_i}), (\mathcal{D}_{u_j}^q, p_{u_j}))],
\end{align*}
where $\phi_{pr}$  is the prompt template for pairwise user preference comparison. The LLM outputs which user has a stronger preference for the query item. To mitigate the position bias in LLM judgment \cite{ye2024justice, lu-etal-2022-fantastically}, for each pair of users, we swap the position of user $u_i$ and $u_j$ and only consider preferences differences if the judgments are consistent across both orderings. Using the pairwise comparison function, we rank users with heapsort, which ensures $O(N\log N)$ computational complexity and has been shown effective in LLM-based text ranking \cite{qin-etal-2024-large}. This process yields the final pairwise ranking $r$.

\paragraph{Listwise Ranking}
Previous research \cite{sun-etal-2023-chatgpt, ma2023zero} has shown that LLMs are effective at listwise text ranking, where they rank the relevance of multiple documents to a query in a single prompt. Similarly, LLMs can rank a group of users within a single prompt input, where each user $u$ is represented by their retrieved rating history $\mathcal{D}_u^q$ and the corresponding user profile $p_u$. The ranking $r$ is defined as:
\begin{align*}
    r = \mathrm{LLM}(\phi_{gp}(q, \{(\mathcal{D}_u^q, p_u), u\in \mathcal{U}^*\})),
\end{align*}
where $\phi_{gp}$ is the prompt construction function for listwise ranking.

\subsection{Evaluation Metric}
\label{sec:metric}
To evaluate LLMs’ personalization capabilities, we measure the correlation between the predicted user preference ranking and the ground truth ranking. The ground truth ranking $r^*$ is derived from the relative ratings within the selected user group:
\begin{align*}
    r^* = \mathrm{argsort}(\{\tilde{y}(q_u), u\in \mathcal{U}^*\}),
\end{align*}
where $\tilde{y}(q_u)$ is the ground truth relative rating. The evaluation metric, \textit{Personalization Tau Correlation (PTC)}, is computed as Kendall’s tau correlation between the predicted ranking $r$ and the ground truth ranking $r^*$ :
\begin{align*}
    \mathrm{PTC} = \textit{Kendall-tau}(r, r^*).
\end{align*}
Overall, we define relative ratings to capture users’ preferences for a query item while eliminating user rating bias. All users within a group are ranked based on the same query item, ensuring consistent item quality. User preference signals are clearly observed through their reviewed ratings, derived from differences in relative ratings across users. By using a single query item, \ourmethod{} expects personalized outputs tailored to each user’s history and profile. This evaluation paradigm is specifically designed to assess personalization capabilities, making the personalization signal easy to interpret.

\section{Experimental Settings}
We evaluate the personalization capabilities of 19 off-the-shelf LLMs, including open-source models: Llama-3.1-8B-it, Llama-3.1-70B-instruct, Meta-Llama-3.1-405B-Instruct \cite{dubey2024llama}, Gemma-2-9B-it, Gemma-2-27B-it \cite{team2024gemma}, Ministral-8B-Instruct-2410, Mistral-Nemo-Instruct-2407, Mixtral-8x22B Instruct v0.1 \cite{jiang2024mixtral}, Qwen2.5-7B-Instruct, Qwen2.5-14B-Instruct, Qwen2.5-32B-Instruct, Qwen2.5-72B-Instruct, Qwen2.5-Coder-32B-Instruct \cite{qwen2024qwen25technicalreport}, DeepSeek-v3 \cite{liu2024deepseek}, and proprietary models: Claude-3.5-haiku, Claude-3.5-sonnet, GPT-4o-mini, and GPT-4o \cite{hurst2024gpt}. For a fair comparison, all models were tested with a temperature of 0.1 using zero-shot prompting by default. For LLM fine-tuning, we applied LoRA \cite{hu2021lora} for efficient fine-tuning with rank 16, training for 2 epochs with a batch size of 32 and a learning rate of  $1\times10^{-5}$. We use BM25 \cite{trotman2014improvements} retriever and the number of retrieved history items $k$ was set to 4 by default, and performance is reported under zero-shot settings without further notice. 

\begin{table*}[t]
  \centering
    \caption{Main results on \ourmethod{}. Scores range from -1 to 1, with higher values indicating better performance. The best results across different LLMs are highlighted in \textbf{bold}, and the second-best results are \underline{underlined}.}
  \begin{adjustbox}{max width=1\linewidth}
    \begin{tabular}{lccccccccc|c}
    \toprule[1.5pt]
    \multirow{2}{*}{\textbf{Model}} & \multicolumn{3}{c}{\textbf{easy}} & \multicolumn{3}{c}{\textbf{medium}} & \multicolumn{3}{c}{\textbf{hard}} & \multirow{2}{*}{\textbf{Avg.}} \\
    \cmidrule(r){2-4} \cmidrule(r){5-7} \cmidrule(r){8-10}
    & pointwise & pairwise & listwise & pointwise & pairwise & listwise & pointwise & pairwise & listwise \\
    \midrule[0.75pt]
    \multicolumn{11}{c}{ {\textbf{\textit{Open-source LLMs}}}}\\
    \hline
    \textsc{Llama3.1-8B-it} & -0.27 & 0.25 & 0.30 & 0.03 & -0.01 & -0.01 & -0.03 & 0.07 & 0.05 & 0.04\\
    \textsc{Gemma-2-9B-it} & 0.13 & 0.23 & 0.25	& 0.00 & 0.10 & -0.08 & 0.03 & 0.05 & 0.09 & 0.09 \\
    \textsc{Qwen2.5-7B-it} & -0.10 & 0.17 & 0.16 & 0.06 & 0.05 & -0.05 & 0.02 & 0.08 & 0.02 & 0.05 \\
    \textsc{Ministral-8B-it} & -0.04	& 0.02 & 0.09 &	0.00 &	0.06& -0.01 &	0.05 & 0.02 & 0.01 & 0.02\\
    \textsc{Mistral-12B-Nemo-it} &-0.11 & 0.14 & 0.37 & 0.00 & 0.07 &	0.01 & 0.05 & 0.02 & -0.10 & 0.05\\
     \textsc{Qwen2.5-14B-it} & 0.09 & 0.23 & 0.17 & 0.05 & 0.10 & 0.03 & 0.08 & 0.09 & 0.02 & 0.10 \\
    \textsc{Gemma-2-27B-it} & 0.21 & 0.15 & 0.15 & 0.02 & 0.03 & -0.01& 0.06 & 0.07 &  0.02 & 0.08 \\
    \textsc{Mixtral-8x22B-it} & -0.02 & 0.25 & 0.35 & 0.05 & 0.11 & 	0.03 & 0.05 & 0.09 & 0.07 & 0.11 \\
    \textsc{Qwen2.5-32B-it} & 0.21 & 0.26 & 0.35 & -0.01 & 0.09 &	0.02 & 0.08 & 0.07 & 0.07 & 0.13 \\
    \textsc{Qwen2.5-Coder-32B-it} & 0.04 & 0.26 & 0.12 & 0.01 & 	0.06 & 0.02 & 0.03 & 0.06 & 0.03 &	0.07\\
    \textsc{Qwen2.5-72B-it} & 0.09 & 0.27 & 0.24 & 0.04 & 0.10 & 0.06 & 0.04 & 0.09 & 0.03 & 0.11 \\
    \textsc{Llama-3.1-70B-it} & 0.09 & 0.31 & 0.32 & 0.00 &  0.09 &	0.13 &	0.09 & 	0.12 & 0.10 & 0.14 \\
    \textsc{Mistral-Large-123B-it} & 0.19 & 0.31 & 0.33 & 0.09 & 0.09 & 0.06 &	0.09 & 0.08	& 0.07 & 0.15 \\
    \textsc{Llama-3.1-405B-it} & 0.14	& 0.31 & 0.38 & 0.08 & 0.09 & 0.05 & 0.05 & 0.08 & 0.11 & 0.14 \\
    \textsc{DeepSeek-v3-671B} & 0.23 & 0.21 & 0.23 & 0.13 & 0.10 & 0.06 & 0.14 & 0.06 & 0.06 & 0.14\\
    \hline
     \multicolumn{11}{c}{\textbf{{\textit{Proprietary LLMs}}}}\\
     \hline
    \textsc{Claude-3.5-Haiku}  & 0.21 & 0.27 & 0.32 & 0.09 & 0.08 & 0.10	& 0.09 & 0.04 & 0.02 & 0.14 \\
    \textsc{Claude-3.5-Sonnet}  & 0.27& 0.34 & 0.31 & 0.11 & 0.07 & 0.14	& 0.13 & 0.10& 0.12 & 0.18\\
    \textsc{GPT-4o-mini}  & 0.20 & 0.31 & 0.41& 0.13 & 0.04 & 0.05 & 0.07	& 0.05& 0.04	& 0.15 \\
    \textsc{GPT-4o}  & 0.25 & 0.29& 0.27 & 0.10 & 0.07 & 0.11 & 0.08 & 	0.07 & 0.06& 0.15\\
    \bottomrule[1.5pt]
    \end{tabular}%
    \end{adjustbox}
   
    \label{main_results}
\end{table*}

\section{Results}
Table \ref{main_results} shows the performance of 19 off-the-shelf LLMs on \ourmethod{}. We have the observations as follows.

\paragraph{LLMs struggle with personalized recommendation.}
Across 19 strong LLMs, performance on \ourmethod{} ranges from 0.02 to 0.18, within Kendall’s tau value range of $[-1, 1]$. This indicates a low to moderate correlation between predictions and ground truth rankings. Pointwise, pairwise, and listwise ranking methods all demonstrate limited success, with average Kendall’s tau scores ranging from -0.27 to 0.38 across models and methods. Even the best-performing model, \textsc{Claude-3.5-Sonnet}, achieves only 0.18 on average across different group sizes and ranking methods. These results highlight the limited personalized preference understanding ability of current LLMs, underscoring that LLM personalized recommendation remains an open research question.

\paragraph{Scaling law does not always hold for personalization.}
While the scaling law suggests larger models generally perform better on tasks \cite{kaplan2020scaling}, our results show that increasing model size does not consistently improve personalization performance. For example, in the \textsc{Qwen} model series, the 7B, 14B, and 32B models perform as expected with scores of 0.05, 0.11, and 0.13, respectively. However, the 72B model performs worse than the 32B model and similarly to the 14B model. Similarly, in the \textsc{Gemma} series, the 27B model performs close to the 9B model. These results challenge the assumption that larger models inherently enhance personalization capabilities.

\paragraph{Pairwise and listwise ranking outperform pointwise.}
Across all user group sizes and models, the average Kendall’s tau scores for pointwise, pairwise, and listwise ranking are 0.19, 0.38, and 0.35, respectively. Pairwise and listwise ranking methods significantly outperform pointwise ranking. We attribute this to the limitations of pointwise ranking, where the model evaluates a single user in isolation, making it difficult to discern subtle preference differences. In contrast, pairwise and listwise methods allow the model to leverage comparative reasoning, capturing nuanced differences by analyzing multiple users within a single prompt.


\paragraph{Strong open-source LLMs rival proprietary models.}
Open-source models demonstrate competitive performance compared to proprietary counterparts on \ourmethod{}. For instance, \textsc{Mistral-Large-123B-it} and \textsc{Llama-3.1-405B-it} achieve average Kendall’s tau scores of 0.15, slightly outperforming \textsc{Claude-3.5-Haiku} and approaching the performance of the \textsc{GPT-4o} family and \textsc{Claude-3.5-Sonnet}. These results suggest that with proper optimization, open-source models can be viable alternatives for personalization tasks, offering performance close to or on par with commercial models.

\section{Analysis}
\begin{figure}
    \centering
    \includegraphics[width=1\linewidth]{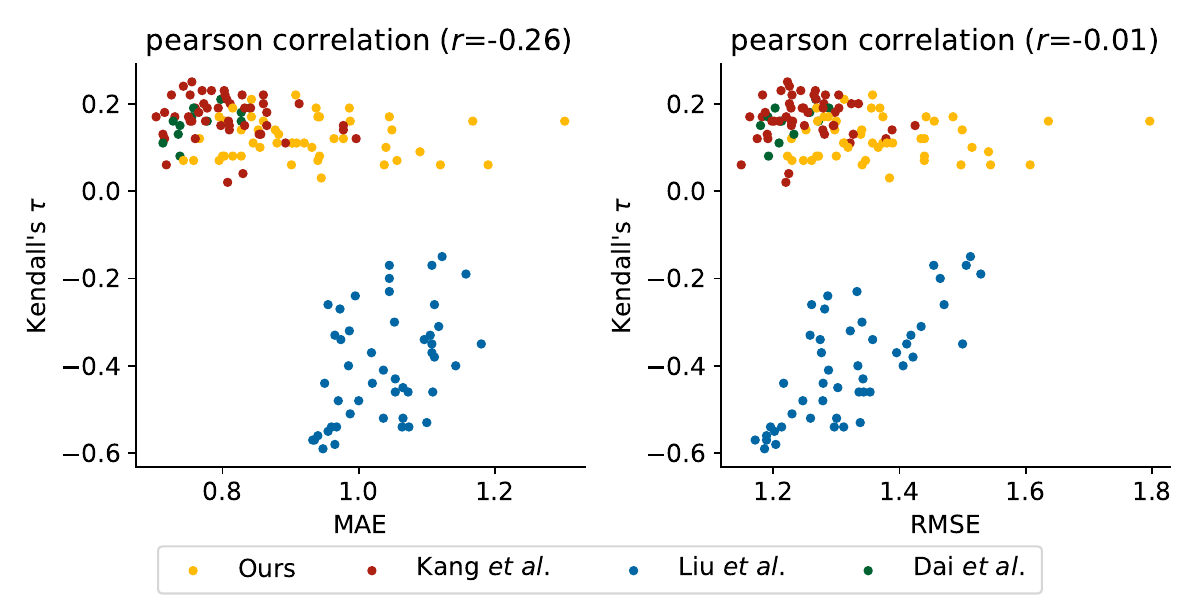}
    \caption{Correlation between Kendall’s tau in \ourmethod{} and traditional metrics (MAE and RMSE). The results show weak correlation, further confirming that MAE and RMSE are not reliable indicators of personalization capabilities.}
    \label{fig:MAE_RMSE_correlation}
\end{figure}

\paragraph{Tau's Correlation with MAE and RMSE}
To validate that \ourmethod{} successfully isolates personalization capabilities in LLM recommendations, while traditional metrics like MAE and RMSE do not, we analyze their correlation with \ourmethod{} performance. Specifically, we adopted prompt templates from \citet{kang2023llms, liu2023chatgpt, dai2023uncovering} and used the default prompting in \ourmethod{}. Additionally, we generated four prompt variants using \textsc{GPT-4o} based on the original prompts.
We varied the number of retrieved history items $k$ in $\{2, 4, 8\}$ and the number of shots in $\{0, 1, 2, 3\}$. For each configuration, we computed MAE, RMSE, and \ourmethod{} performance, setting the decoding temperature to 0 to eliminate randomness. The correlations between Kendall’s tau, MAE, and RMSE are visualized in Figure \ref{fig:MAE_RMSE_correlation}, alongside the corresponding Pearson correlation coefficients.
The results show that both MAE and RMSE have weak correlations with Kendall’s tau in \ourmethod{}. Notably, while \citet{liu2023chatgpt} demonstrates moderate performance on MAE and RMSE, its performance on \ourmethod{} consistently falls below random guessing. This indicates that traditional rating prediction metrics like MAE and RMSE are poor indicators of personalization capabilities.


\paragraph{Comparing Prompting Methods}
Previous studies \cite{richardson2023integrating, tan2024democratizing} have shown that incorporating textual user profiles into prompts can enhance model performance. However, conflicting conclusions exist regarding whether few-shot prompting improves LLM personalization capabilities \cite{kang2023llms, zhiyuli2023bookgpt}. To address this, we evaluate personalization performance under different prompting methods using \textsc{GPT-4o-mini}, with results presented in Table \ref{tab:prompt_format}. Including user profiles $p_u$ in prompts leads to an average performance improvement of 28\%. Few-shot prompting, while beneficial for pointwise and listwise methods, reduces performance in pairwise prompting, resulting in relatively stable overall performance. Combining user profiles with few-shot prompting yields worse performance than zero-shot prompting with user profiles. Self-consistency prompting (with 5 sampling times) improved performance in simpler ranking tasks but showed negligible or no gains in more challenging settings. Chain-of-thought prompting, which guides relative rating computation and ranking steps, unexpectedly degraded performance, particularly for easy tasks. These findings suggest that prompting strategies such as few-shot, self-consistency, and chain-of-thought are not universally effective for enhancing personalization performance. This may be due to inconsistencies between user behavior patterns in few-shot demonstrations and those in the query task. These results validate the design choice of profile-augmented zero-shot prompting as the primary method in our main experiments, as it strikes a better balance between simplicity and performance.

\begin{table}[t]
    \centering
    \caption{Performance of \textsc{GPT-4o-mini} on \ourmethod{} with different prompting methods, where the best performance across prompting method is in \textbf{bold}, the second best is \underline{underlined}. Incorporating user profiles significantly enhances personalization capabilities, whereas few-shot, self-consistency, and chain-of-thought prompting does not consistently improve performance and may even degrade it.}
    \resizebox{1\linewidth}{!}{
    \begin{tabular}{l c c c c c c c c c c c }
        \toprule[1.5pt] 
        \multirow{2}{*}{\textbf{Prompting}} & \multicolumn{3}{c}{\makecell[c]{\textbf{easy}}} & \multicolumn{3}{c}{\makecell[c]{\textbf{medium}}} & \multicolumn{3}{c}{\makecell[c]{\textbf{hard}}} &\multirow{2}{*}{\textbf{Avg.}} \\          
         \cmidrule(r){2-4} \cmidrule(r){5-7} \cmidrule(r){8-10}   
          & pt & pr & ls & pt & pr & ls & pt & pr & ls & \\
        \cmidrule(r){1-11}
        \textsc{Zero-shot} & .20 & .31 & .41 & .13 & .04 & .05 & .07 &  .05 & .04 & .15\\
        \textsc{Zero-shot w/o profile} & .14 & .24 & .27 & .11 & .06 & .02 & .05 & .08 & .05 & .11\\
          \textsc{Few-shot} & .15 & .31 & .37 & .13 & .07 & .03 & .09 &-.02 & .06 & .13\\
         \textsc{Few-shot w/o profile} & .16 & .20 & .29 & .10 & .07 &  .03 &  .08 & .03 &	.08 & .11\\
         \textsc{Self-Consistency} & .28 & .33 & .35 & .11 & .05 & .04 & .07 & .05 & .03 & .15 \\
         \textsc{Chain-of-Thought} & .09 & .26 & .31 & .08 & .08 & .08 & .08 & .05 & .01 & .12\\

       \bottomrule[1.5pt]
    \end{tabular}
    }
    
    \label{tab:prompt_format}
\end{table}

\begin{table*}[t]
  \centering
    \caption{\ourmethod{} results of \textsc{Llama-3.1-8B-it} and \textsc{Mistral-12B-Nemo-it} with different supervised fine-tuning strategies, where the best results across different SFT methods is in \textbf{bold}, and the second best is \underline{underlined}. While the weight merging generally achieves the best performance, it fails to achieve universal improvement across different ranking methods and task difficulties.}
  \begin{adjustbox}{max width=1\linewidth}
    \begin{tabular}{l ccccccccc |c}
    \toprule[1.5pt]
    \multirow{2}{*}{\textbf{Training Method}} & \multicolumn{3}{c}{\textbf{easy}} & \multicolumn{3}{c}{\textbf{medium}} & \multicolumn{3}{c}{\textbf{hard}} & \multirow{2}{*}{\textbf{Avg.}} \\
    \cmidrule(r){2-4} \cmidrule(r){5-7} \cmidrule(r){8-10} 
       & pointwise & pairwise & listwise & pointwise & pairwise & listwise & pointwise & pairwise & listwise\\
    \midrule[0.75pt]
    \multicolumn{11}{c}{\ \  \textbf{\textsc{Llama-3.1-8B-it}}}\\
    \hline
     \textsc{Prompting} & -0.27 & 0.25 & 0.30 & 0.03 & -0.01 & -0.01 & -0.03 & 0.07 & 0.05 & 0.04\\
    \textsc{Pointwise Only} & 0.13 & 0.21 & - & 0.18 & 0.00 & - & 0.11 & 0.03 & - & -\\
    \textsc{Pairwise Only} & -0.04 & 0.21 & - & 0.18 & 0.06 & - & 0.18 & 0.09 & - & -\\
    \textsc{Listwise Only} & -0.14 & 0.15 & -0.05 & 0.18 & 0.03 & -0.03 & 0.19 & 0.03 & 0.05 & 0.04\\
    \textsc{Multi-task Training} & 0.12 & 0.00 & 0.04 & 0.16 & -0.02 & -0.01 & 0.13 & -0.01 & 0.04 & 0.05\\
    \textsc{Weight Merging} & 0.15 & 0.22 & 0.31 & 0.22 & -0.01 & -0.03 & 0.16 & 0.08 & 0.06 & 0.13\\
    \midrule
    \multicolumn{11}{c}{ \textbf{\textsc{Mistral-12B-Nemo-it}}}\\
    \hline
    \textsc{Prompting} & -0.11 & 0.14 & 0.37 & 0.00 & 0.07 & 0.07 & 0.05 & 0.02 & -0.10 & 0.07 \\
    \textsc{Pointwise Only} & 0.17 & 0.31 & 0.20 & 0.07 & -0.01 & 0.09 & 0.08 & 0.02 & 0.01 & 0.12\\
    \textsc{Pairwise Only} & 0.17 & 0.40 & 0.26 & 0.02 & -0.03 & 0.06 & 0.05 & 0.03 & 0.06 & 0.12\\
    \textsc{Listwise Only} & 0.17 & 0.30 & 0.22 &  0.10 & 0.08 & 0.05 & 0.08 & -0.01 & -0.02 & 0.11\\
    \textsc{Multi-task Training} & 0.17 & 0.11 & 0.12 & 0.16 & 0.01 & 0.02 & 0.11 & 0.01 & 0.03 & 0.08 \\
    \textsc{Weight Merging} & 0.19 & 0.34 & 0.21 & 0.08 & 0.08 & 0.07 & 0.06 & 0.01 & 0.11 & 0.13 \\

    \bottomrule[1.5pt]
    \end{tabular}%
    \end{adjustbox}
   
    \label{improvement}
\end{table*}

\begin{figure}[t]
    \centering
    \includegraphics[width=1\linewidth]{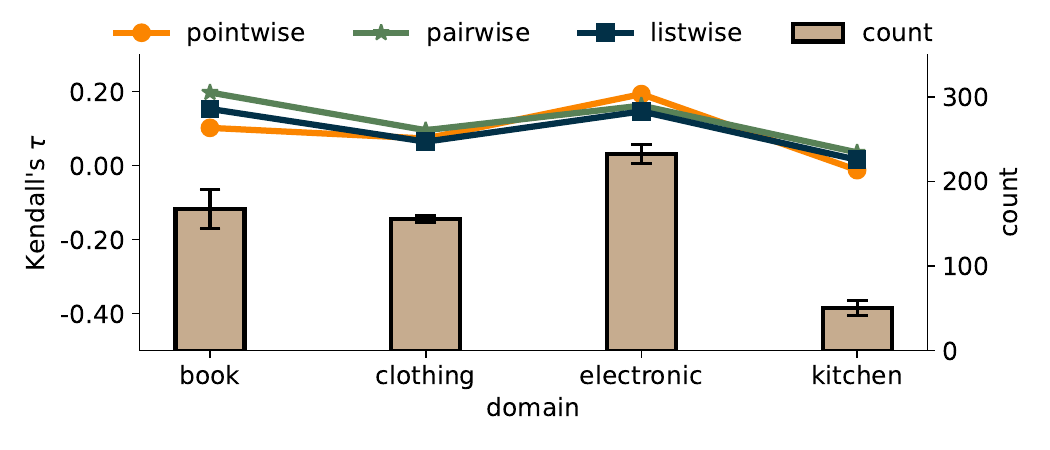}
    \caption{Performance across different domains and average item name count in the pretraining dataset. Query items with higher frequency in pretraining data generally show better performance in \ourmethod{}.}
    \vspace{-1.0cm}
    \label{fig:domain}    
\end{figure}

\paragraph{Performance across Different Domains}
Personalized recommendations span various domains, each with unique data distributions that can influence model performance. To examine this, we analyzed the correlation between model performance across different shopping domains and the frequency of item names in the LLM pretraining dataset. Specifically, we selected 50 user groups from the book, clothing, electronic, and kitchen domains, comparing their performance on \ourmethod{} with the average frequency of item names in the pretraining corpus.
Since the training corpora of base LLMs are not publicly accessible, we approximated using the Dolma Corpus \cite{soldaini2024dolma}, which contains 3.1T tokens. The infini-gram method \cite{liu2024infini} was used to calculate the average occurrence of item names in the corpus. As shown in Figure \ref{fig:domain}, query items with higher frequencies in the pretraining dataset generally exhibit better performance in \ourmethod{}.
These findings suggest that LLM personalization capabilities are partially influenced by the domain distribution of their pretraining data, emphasizing the importance of diverse training datasets for improving performance across domains.

\section{Enhancing LLM Personalized Recommendation}
Results from \ourmethod{} reveal that current LLMs are not effective personalized recommender systems, as they show limited capability in understanding user preferences when user rating bias and item quality are eliminated. To address this limitation, we explore several supervised fine-tuning (SFT) strategies using held-out user groups. The methods are as follows:
\begin{itemize}[leftmargin=*, itemsep=0pt, topsep=2pt, parsep=0pt,partopsep=0pt]
    \item \textbf{Single-task Training} This approach involves training the model exclusively on one ranking method from pointwise, pairwise, or listwise.
    \item \textbf{Multi-task Training} In this method, pointwise, pairwise, and listwise ranking tasks are combined to create a joint training dataset for SFT.
    \item \textbf{Weight Merging} Parameters trained separately on pointwise $\theta_{pt}$, pairwise $\theta_{pr}$, and listwise methods $\theta_{ls}$ are merged using a linear combination: $\theta_{fnl}=\alpha \times \theta_{pt} + \beta \times\theta_{pr} + (1-\alpha - \beta)\times \theta_{ls}$. We set $\alpha=\beta=1/3$, averaging the weights from the three models. 
\end{itemize}

\paragraph{Results} We evaluate these SFT methods using \textsc{Llama-3.1-8B-it} and \textsc{Mistral-12B-Nemo-it}, with results shown in Table \ref{improvement}. Surprisingly, single-task training not only improves performance in the targeted ranking task but also shows moderate improvement in other tasks, suggesting effective task transfer. Conversely, multi-task training often underperforms compared to single-task training, indicating potential negative task transfer.
The weight merging method consistently delivers the best results, achieving a notable improvement in average Kendall’s tau and performance close to that of 70B parameter models. These findings highlight the importance of positive task transfer for developing LLMs capable of pointwise, pairwise, and listwise preference ranking. Though cannot achieve universal improvement against direct prompting, weight merging emerges as a viable improve strategy.



\section{Related Work}

\paragraph{Evaluation Metric of Personalized Recommendation}

Personalized recommendation systems are evaluated using metrics tailored to ranking-based and rating-based tasks. For ranking-based recommendations, models predict an ordered list of items, evaluated using metrics such as Normalized Discounted Cumulative Gain (NDCG) \cite{jarvelin2002cumulated} and Hit Rate (HR) for ranking quality. Mean Reciprocal Rank (MRR) \cite{radev2002evaluating} assesses the position of the first relevant item, while Precision@K and Recall@K evaluate the relevance of top-k recommendations. Metrics like Coverage and Diversity capture the range of items recommended and their dissimilarity, respectively.
For rating-based tasks, where models predict user ratings for query items. Metrics such as Mean Absolute Error (MAE) and Root Mean Squared Error (RMSE) \cite{willmott2005advantages} measure regression error, while R-Squared \cite{nagelkerke1991note} evaluates the fit between predictions and actual ratings. Rating prediction can also be treated as a binary classification task, with metrics like AUC-ROC \cite{hanley1982meaning}, F1 Score, Precision, Recall, and Log Loss used to assess performance. 

\paragraph{LLM-based Personalized Recommendation}

Existing LLM-based methods for personalized recommendations can be broadly classified into ranking-based and rating-based approaches.
Ranking-based methods generate an ordered list of items for users, leveraging user behavior history to predict the top-K items of interest. In-context learning is a popular paradigm, with several works exploring exemplars to improve understanding of user preferences \cite{liu2023chatgpt, dai2023uncovering, zhang2023chatgpt, liu2023genre, hou2023large, du2023enhancing}. \citet{zhang2023recommendation} introduced Chain-of-Thought prompting for top-K recommendations. Fine-tuning LLMs has also been explored to improve representation and domain adaptation. For instance, \citet{chen2023palr} and GenRec \cite{ji2023genrec} fine-tuned Llama-7B for recommendation tasks, while \citet{zhang2023recommendation} adapted Flan-T5-XL through instruction tuning. \citet{harte2023leveraging} integrated LLM embeddings and prompts with traditional sequential recommendation approaches.
Rating-based methods predict user ratings for specific items, probing LLMs’ ability to understand preferences and predict user behavior. Similar to ranking tasks, these methods can involve frozen or fine-tuned LLMs. For frozen LLMs, BookGPT \cite{zhiyuli2023bookgpt} and \citet{dai2023uncovering} used prompt engineering for rating prediction, while KAR \cite{xi2023towards} generated user profiles and item knowledge for use in discriminative recommendation systems. Fine-tuned LLMs include \citet{kang2023llms}, which adapted LLMs to rating prediction tasks, and TallRec \cite{bao2023tallrec}, which integrated LoRA and instruction tuning. OPPU \cite{tan2024democratizing} introduced personalized PEFT for private and accurate rating prediction, while Per-Pcs \cite{tan2024personalized} optimized PEFT for efficient personalization.

Ranking-based evaluations are limited by their reliance on unreviewed items as distractors, as these items may still be suitable recommendations despite lacking user exposure. Rating-based methods are influenced by user rating biases and item quality, often failing to accurately reflect personalization capabilities. To address these limitations, we propose \ourmethod{}, a benchmark that eliminates biases from ratings and item quality, relying solely on observed user preference signals to evaluate true personalization effectiveness.

\section{Conclusion}
User rating bias and item quality significantly impact user ratings, often hindering the evaluation of user preferences. To address this, we introduced \ourmethod{}, a personalized recommendation benchmark that removes the influence of rating bias and item quality, focusing solely on observed user preference signals to evaluate personalization capabilities. Extensive experiments on \ourmethod{} reveal that current LLMs face substantial challenges in personalization, with performance varying based on prompting methods and training domains. We also investigated supervised fine-tuning strategies, finding weight merging to be the most effective, but enhancing LLMs’ personalization capabilities remains an open challenge.



\section*{Limitations}
We identify two key limitations in \ourmethod{}. First, the dataset scale is relatively small, comprising 200 user groups for each difficulty level, resulting in a total of 600 groups and 1800 users. This limitation arises from the strict constraints in selecting personalization data. Additionally, the user group selection criteria in \ourmethod{} may bias the dataset toward more popular items, potentially introducing item-related bias into the evaluation process.
Second, the methods explored for improving LLM personalization capabilities are not universally effective in all scenarios, leaving the challenge of enhancing personalized LLM recommendations an open research problem. Future work could explore encoding user histories into personalized PEFT parameters \cite{tan2024democratizing}, which may offer a promising direction with sufficient computational resources. Furthermore, our experiments employ LoRA for SFT on Llama-3.1-8B-it and Mistral-12B-Nemo-it. While efficient, this approach might impact results compared to full-parameter fine-tuning.

\section*{Ethical Considerations}
\paragraph{Data Bias}
The design of \ourmethod{} relies on observed user preferences, and while efforts are made to eliminate rating bias and item quality effects, biases inherent in the underlying data may still influence the evaluation and personalization capabilities. For instance, popular items may disproportionately appear in user groups, potentially introducing item-related biases. Such biases could skew evaluations and lead to misleading conclusions about LLM personalization performance. Future work should explore methods to ensure diversity and fairness in data selection and mitigate biases in both user and item distributions.

\paragraph{Privacy}
Personalization inherently requires the use of user-specific data, which may include sensitive or private information. While \ourmethod{} focuses on observed user preferences and anonymized data, extending this benchmark to real-world applications may involve privacy risks. Care must be taken to ensure that data used for personalization is anonymized, securely stored, and handled in compliance with privacy regulations. Future iterations of \ourmethod{} could incorporate privacy-preserving techniques, such as differential privacy or personalized parameter-efficient fine-tuning (PEFT), to enhance privacy safeguards.

\paragraph{Accessibility}
The computational demands of LLM training and evaluation, particularly for benchmarks like \ourmethod{}, pose challenges for smaller organizations or individual researchers with limited resources. This may exacerbate disparities in access to cutting-edge personalization research and hinder equitable advancements in the field. Efforts should focus on improving the efficiency of benchmarking frameworks and exploring lightweight alternatives to support broader accessibility and inclusivity in AI research.

\paragraph{Fairness in Personalization}
While \ourmethod{} aims to evaluate personalization capabilities, care must be taken to ensure that such personalization does not inadvertently reinforce harmful stereotypes or exclude certain user groups. Models evaluated on \ourmethod{} should be assessed not only for their personalization accuracy but also for fairness and inclusivity, ensuring equitable treatment across diverse user populations.

\bibliography{custom}
\clearpage
\appendix

\begin{figure}[t]
    \centering
    \includegraphics[width=1\linewidth]{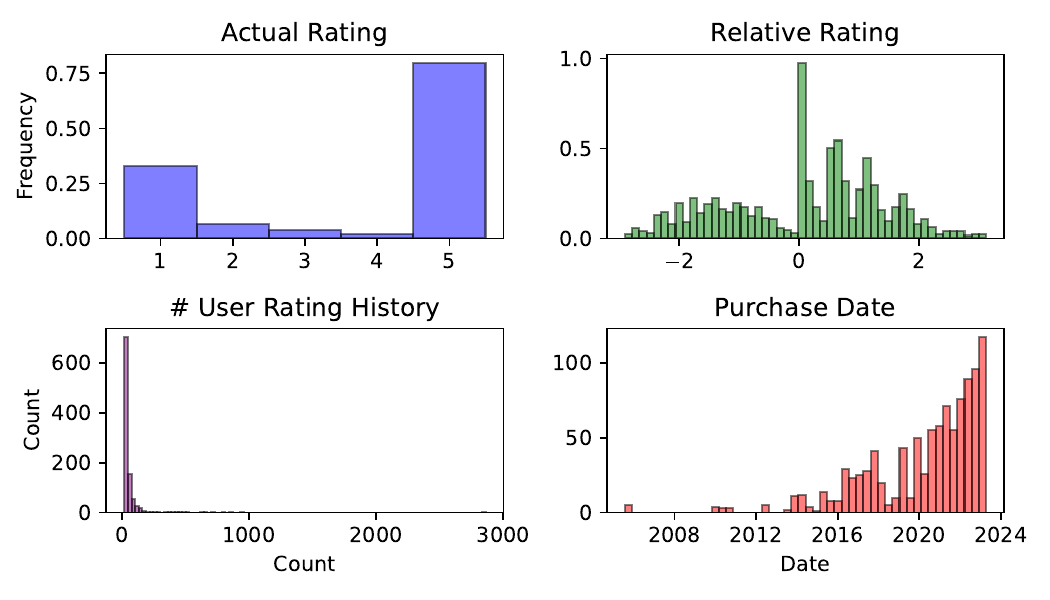}
    \caption{The statistics of \ourmethod{}, including the distribution of actual rating, relative rating, length of user history, and the purchase date.}
    \label{fig:statistics}
\end{figure}

\begin{table}[t]
    \centering
    \caption{Statistics of \ourmethod{}.}
    \resizebox{1\linewidth}{!}{
    \begin{tabular}{l c c}
        \toprule[1.5pt] 
         \textbf{Statistics} & \textbf{Held-out Data} & \textbf{Test Data}\\
        \midrule[0.75pt] 
        \# Review & 2,438 & 1,007\\
        \# User & 2,140 & 986 \\
        \# Query Item & 466 & 200 \\
        \# Rating History & 157,480 & 61,412 \\
        \# Token & 289,164 & 111,017\\
        Domain & \multicolumn{2}{c}{\makecell[c]{book, clothing, electronic, kitchen}}\\
        Time Range & 07/16/2022 - 05/16/2023 & 07/25/2005 - 04/14/2023 \\
       \bottomrule[1.5pt]
    \end{tabular}
    }
    
    \label{tab:stat}
\end{table}

\section{\ourmethod{} Statistics}
We present the statistics of \ourmethod{} in Figure \ref{fig:statistics}, which include the distributions of actual ratings, relative ratings, the number of user rating histories, and the purchase dates of query items. The actual ratings are biased toward scores of 1 and 5. The relative rating distribution peaks around 0 and follows an approximately normal distribution. For the distribution of user rating histories, a long-tail pattern is observed, with all selected users having more than 20 rating histories to ensure sufficient data for personalization. The purchase date distribution shows that most user activity in \ourmethod{} occurred around 2022, indicating that \ourmethod{} contains up-to-date data. Additional benchmark statistics are provided in Table \ref{tab:stat}.

\section{Analysis (Cont.)}

\begin{figure}[t]
    \centering
    \includegraphics[width=1\linewidth]{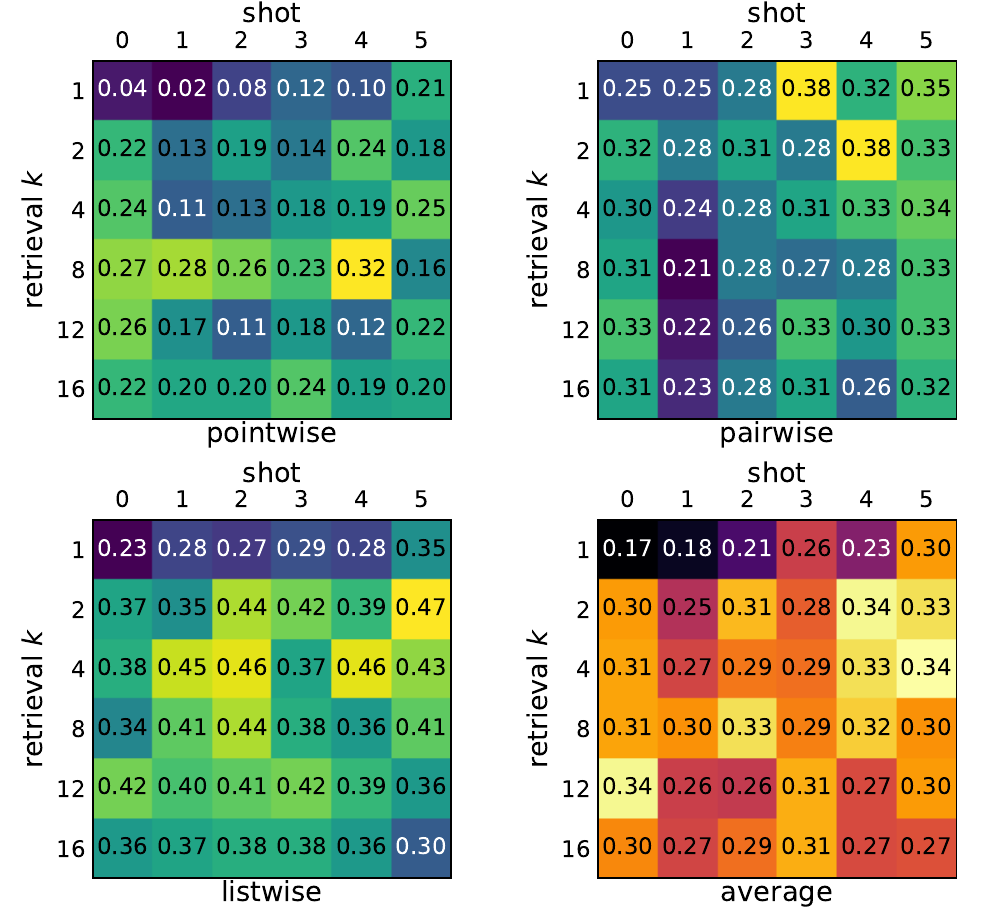}
    \caption{Performance of \ourmethod{}\textsc{-easy} with different retrieval $k$ and and shot number. By searching the appropriate $k$ and shot combination, we can improve the performance.}
    \label{fig:shot_retrieval}
\end{figure}

\paragraph{Combination of Shot and $k$}
Although the preliminary results in Table \ref{tab:prompt_format} indicate that few-shot prompting does not consistently enhance personalization capabilities, its interaction with different retrieval history counts ($k$) warrants further investigation to understand the role of demonstrations in LLM personalization. To explore this, we vary the shot size ($\text{shot} \in \{0, 1, 2, 3, 4, 5\}$) and the retrieval history count ($k \in \{1, 2, 4, 8, 12, 16\}$) and visualize the performance for pointwise, pairwise, and listwise evaluations, as well as the average performance under the easy setting in Figure \ref{fig:shot_retrieval}. The results show that while few-shot prompting does not consistently improve performance and can sometimes degrade it, it provides relatively consistent improvements when $k \in \{2, 4\}$ and $\text{shot} \in \{4, 5\}$. Furthermore, when $k = 1$, increasing the number of shots results in noticeable improvements; however, this benefit diminishes as $k$ increases. We hypothesize that while both shot examples and retrieval histories provide relevant information, the retrieval history is more directly aligned with the user’s personalized preferences. In contrast, patterns in randomly chosen shots may introduce noise, distracting the LLM and negatively affecting predictions.


\begin{figure*}[t]
    \centering
    \includegraphics[width=1\linewidth]{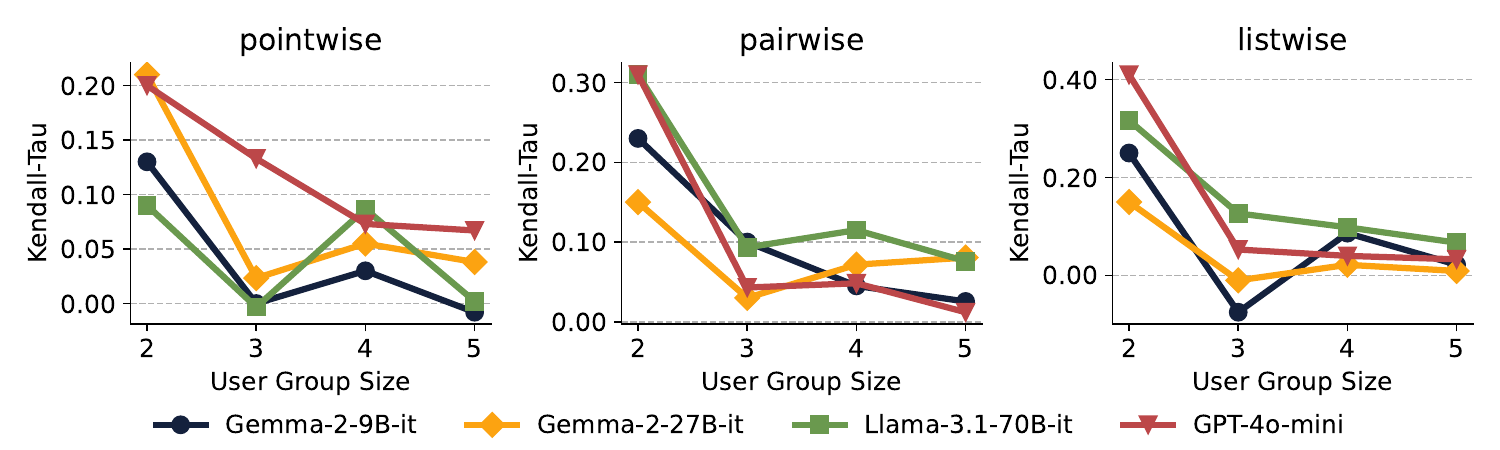}
    \caption{LLM performance using pointwise, pairwise, and listwise ranking methods across different user group sizes. Larger user groups generally result in reduced performance, while pointwise ranking demonstrates stroner robustness to group size variations.}
    \label{fig:groupsize}
\end{figure*}

\paragraph{Performance \emph{w.r.t.} User Group Size}
From the main results, the average performance for user group sizes of 2, 3, and 4 is 0.201, 0.052, and 0.058, respectively. These findings suggest that groups with two users are less challenging for LLMs, as they only require a single comparison between the two users. However, as the group size increases beyond two, the number of comparisons required for user preference evaluation grows, leading to significant performance degradation.
We visualize the performance of pointwise, pairwise, and listwise rankings across different user group sizes in Figure \ref{fig:groupsize}. The results show that pointwise ranking is more robust to variations in group size, with performance decreasing only slightly from 0.09 to 0.05 and 0.06. In contrast, pairwise and listwise rankings exhibit significant drops when the group size exceeds two, with pairwise ranking declining from 0.24 to 0.07 and listwise ranking from 0.27 to 0.05. This trend may be attributed to the accumulation of errors in pairwise ranking and the increased complexity of the task in listwise ranking as the group size grows.


\begin{figure}[t]
    \centering
    \includegraphics[width=1\linewidth]{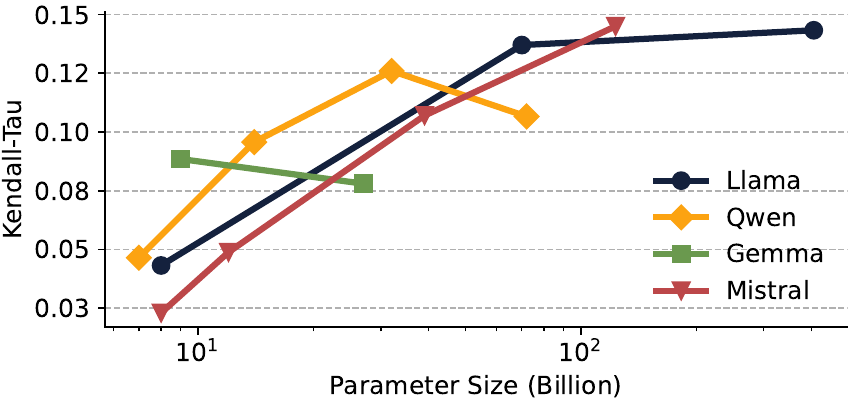}
    \caption{Zero-shot average performance across varying task difficulties and ranking methods for models of different parameter sizes in \ourmethod{}. The Llama, and Mistral model families demonstrate performance improvements with increased model size, while Gemma and Qwen family cannot guarantee the performance gain with larger model size.}
    \label{fig:scaling}
\end{figure}

\paragraph{Scaling Law in \ourmethod{}}
We analyze the performance of the Llama, Qwen, Gemma, and Mistral model families to investigate the scaling law in \ourmethod{}, as shown in Figure \ref{fig:scaling}. For the Llama and Mistral families, larger parameter sizes consistently lead to better performance, aligning with the scaling law. However, this trend does not hold uniformly for the Qwen and Gemma model families. For instance, the Qwen-2.5-72B-it model significantly underperforms the Qwen-2.5-32B-it model, and the Gemma-2-27B-it model falls short compared to the Gemma-2-9B-it model. These findings suggest that while larger models often exhibit stronger performance in \ourmethod{}, simply increasing parameter size does not guarantee performance improvements. This challenges the universal applicability of the scaling law in \ourmethod{}, indicating that factors beyond parameter count may play a critical role in model performance.

 


\section{Computational Resources}
All experiments are implemented on a server with 3 NVIDIA A6000 GPU and Intel(R) Xeon(R) Silver 4210R CPU @ 2.40GHz with 20 CPU cores. For model fine-tuning, it takes 9 hours, 4 hours, 12 hours, and 25 hours to do pointwise, pairwise, and listwise supervised fine-tuning on a single GPU.

\section{Scientific Artifacts}
\ourmethod{} is built with the help of many existing scientific artifacts, including PyTorch \cite{paszke2019pytorch}, Numpy \cite{harris2020array}, huggingface, vllm \cite{kwon2023efficient}, and transformers \cite{wolf2020transformers}. 
We will make the \ourmethod{} data, source code, and all model output publicly available to facilitate further research.

\section{Prompt Template}
\begin{prompt}{Pointwise Prompt Template}
[System]\\
Act as a personalized product recommender system. Below is a list of user's rating history, shown in [User History]. Your task is to predict the user's rating for the query item, which is described in [Query Item Details]. Analyze each user's preferences for the query item based on their historical ratings to generate the prediction. Output a predicted rating ranging from 1 to 5, where 1 being not recommended and 5 being highly recommended. The final answer should strictly follow this JSON structure: \{"predicted\_rating": <rating>\}\\

[User History]\\
\{\texttt{User Data}\}\\

[Query Item Details]\\
\{\texttt{Query Item Information}\}
\\

Answer:
\end{prompt}

\begin{prompt}{Pairwise Prompt Template}
[System]\\
Act as a personalized product recommender system. Below is a pair of users, [User A] and [User B], each with their rating history. Your task is to determine which user is more likely to prefer the query item, based on its details in [Query Item Details]. Analyze each user’s preferences for the query item using their historical ratings and output the user more likely to prefer the query item.

[User A]\\
\{\texttt{User A Information}\}\\

[User B]\\
\{\texttt{User B Information}\}\\

[Query Item Details]\\
\{\texttt{Query Item Information}\}

Which user prefer the query item more? Output only "[User A]" or "[User B]", do not generate anything else:
\end{prompt}

\begin{prompt}{Listwise Prompt Template}
[System]\\
Act as a personalized product recommender system. Below is a group of users accompanied with each user's rating history, shown in [Users]. Your task is to rank these users based on their preference of the query item, which is described in [Query Item Details]. Analyze each user's preferences for the query item based on their historical ratings to generate this ranking. Output the list of user indices (e.g., 1 for User1), ranked from highest to lowest preference for the query item. The final output should rank users from most preferred to least preferred for the query item and adhere to the JSON structure shown below: \{"predicted\_ranking": <user\_ranking>\}\\

[Users]\\
\{\texttt{All User Information within User Group}\}\\

[Query Item Details]\\
\{\texttt{Query Item Information}\}\\

The answer ranks users from most preferred to least preferred for the query item and adhere to the following JSON format, output the list of user indices (e.g., 1 for User1), do not include any additional information: \{"predicted\_ranking": <user\_ranking>\}\\
Answer:
\end{prompt}

\begin{prompt}{Prompt Template for User Profile Generation}
\#\#\# User Behavior history\\
\{\texttt{User Behavior List}\}\\

\#\#\# Task Instruction\\
You are given a list of user behavior history data. Your task is to analyze this data and create a user profile that describes the user's preferences, interests, and patterns of behavior. This profile should be written in a concise and coherent narrative form. Only generate user profile without any additional characters or formatting. 
    
\end{prompt}

\begin{prompt}{Template for a Single Behavior}
\#\#\# Item Title\\
\{\texttt{Item Name}\}\\

\#\#\# Item Author\\
\{\texttt{Author Name}\}\\

\#\#\# User Rating\\
\{\texttt{User Rating}\}
\end{prompt}

\begin{prompt}{Template for a Single User}
<|The Start of User Data|>\\
\#\#\# User Profile\\
\{\texttt{User Profile}\}\\

\#\#\# User Most Common Rating\\
\{\texttt{User Most Common Rating}\}\\

\#\#\# User Average Rating\\
\{\texttt{User Average Rating}\}\\

\{\texttt{Retrieved Top-k User History Behavior}\}\\

<|The End of User Data|>
\end{prompt}

\begin{prompt}{Query Item Template}
<|The Start of Query Item Information|>\\
\#\#\# Item Title\\
\{\texttt{Item Name}\}\\

\#\#\# Item Author\\
\{\texttt{Item Author}\}\\
<|The End of Query Item Information|>
\end{prompt}

\begin{prompt}{Pointwise Chain-of-Thought Prompt Template}
[System]\\
To predict a user’s rating for a query item, follow these steps:
1. Analyze the user’s preference for the query item using their history and profile.
2. Predict the user’s rating for the query item: If the item is likely preferred by the user, the predicted rating should be higher than the user’s average rating. If the item is unlikely to be preferred by the user, the predicted rating should be lower than the user’s average rating.
Act as a personalized product recommender system. Below is a list of user's rating history, shown in [User History]. Your task is to predict the user's rating for the query item, which is described in [Query Item Details]. Analyze each user's preferences for the query item based on their historical ratings to generate the prediction. Output a predicted rating ranging from 1 to 5, where 1 being not recommended and 5 being highly recommended. The final answer must strictly follow this JSON structure: \{"predicted\_rating": <rating>\}.\\

[User History]\\
\{\texttt{User Data}\}\\

[Query Item Details]\\
\{\texttt{Query Item Information}\}\\

Answer: Let's think step by step.
\end{prompt}

\begin{prompt}{Pairwise Chain-of-Thought Prompt Template}
[System]\\
Act as a personalized product recommender system. Below is a pair of users, [User A] and [User B], each with their rating history. Your task is to determine which user is more likely to prefer the query item, based on its details in [Query Item Details]. Analyze each user’s preferences for the query item using their historical ratings and output the user more likely to prefer the query item.\\

[User A]\\
\{\texttt{User A Information}\}\\

[User B]\\
\{\texttt{User B Information}\}\\

[Query Item Details]\\
\{\texttt{Query Item Information}\}\\

Which user prefer the query item more? First provide a short thinking step, then output your final answer in "[User A]" or "[User B]".\\
Answer: 
\end{prompt}

\begin{prompt}{Listwise Chain-of-Thought Prompt Template}
[System]\\
To rank user preferences, follow these steps:
1.	Predict the rating of the query item for each user.
2.	Calculate the relative rating by subtracting each user’s average rating from the predicted rating.
3.	Rank user preferences based on the relative ratings: users with higher relative ratings should be ranked higher, while those with lower relative ratings should be ranked lower.
Act as a personalized product recommender system. Below is a group of users accompanied with each user's rating history, shown in [Users]. Your task is to rank these users based on their preference of the query item, which is described in [Query Item Details]. Analyze each user's preferences for the query item based on their historical ratings to generate this ranking. Output the list of user indices (e.g., 1 for User1), ranked from highest to lowest preference for the query item. The final output should rank users from most preferred to least preferred for the query item and adhere to the JSON structure shown below: \{"predicted\_ranking": <user\_ranking>\}\\

[Users]\\
\{\texttt{All User Information within User Group}\}\\

[Query Item Details]\\
\{\texttt{Query Item Information}\}\\

The answer ranks users from most preferred to least preferred for the query item and adhere to the following JSON format, output the list of user indices (e.g., 1 for User1): \{"predicted\_ranking": <user\_ranking>\}\\
Answer: Let's think step by step.
\end{prompt}



\begin{small}


\begin{table*}[t]
\caption{Pointwise ranking output in \ourmethod{}.}
\begin{adjustbox}{max width=1\linewidth}
\begin{tabular}{p{2in}p{5in}}
    \toprule[1.5pt]
    \textbf{Input} & [User History]\newline<|The Start of User Data|>\newline\#\# User Profile\newline
    The user appears to have a strong interest in home decor and organization. They have purchased various items such as wall art, a headboard, a folding table, bed sheets, and storage ottomans, suggesting a desire to create a comfortable and stylish living environment. Additionally, their purchase of a vacuum cleaner and coat rack indicates a concern for cleanliness and organization.\newline The user also seems to have a preference for quality and durability, as evidenced by their purchase of high-thread-count sheets and a stainless steel coffee percolator. They may value products that are long-lasting and well-made.\newline Furthermore, the user has shown an interest in area rugs, with purchases ranging from small runners to larger rugs, potentially indicating a desire to add warmth and texture to their living spaces.\newline Overall, the user profile suggests an individual who values a well-organized and aesthetically pleasing home environment, with a focus on quality and functional pieces that contribute to both comfort and style.\newline\#\# User Most Common Rating\newline5.0\newline\#\# User Average Rating\newline3.9\newline\#\#\# Item Title\newline West Bend 54159 Classic Stainless Steel Electric Coffee Percolator with Heat Resistant Handle and Base Features Detachable Cord, 12-cup, Silver\newline\#\#\# User Rating\newline2.0\newline\#\#\# Item Title\newline Cloth Napkins Set of 12 Cotton Linen Blend Printed Dinner Napkins Perfect for Parties Dinners Weddings Cocktail Christmas Napkins Cloth 20x20 Blue Floral\newline\#\#\# User Rating\newline5.0\newline\#\#\# Item Title\newline Weavric Cloth Dinner Napkin Bulk, Set of 12, 20 X 20 Inches Wrinkle-Free Washable Reusable Forest Green Linen Table Napkins with Hemmed Edge for Wedding, Party, Hotel, Restaurant\newline\#\#\# User Rating\newline5.0\newline\#\#\# Item Title\newline Crown Mark Barlow Bicast Headboard, King\newline\#\#\# User Rating\newline5.0\newline<|The End of User Data|>\newline\newline[Query Item Details]\newline<|The Start of Query Item Information|>\newline\#\#\# Item Title\newline Keurig K-Slim Coffee Maker, Single Serve K-Cup Pod Coffee Brewer, Multistream Technology, Scarlet Red\newline<|The End of Query Item Information|>\newline Answer: \\
    \hline 
    \textbf{Gemma-2-9B-it} & \{"predicted\_rating": 4\}\\
    \hline
    \textbf{Qwen-2.5-14B-it} & \{"predicted\_rating": 4\} \\
    \hline 
    \textbf{Llama-3.1-70B-it}& \{"predicted\_rating": 2.0\}\\
    \hline
    \textbf{GPT-4o-mini} & \{"predicted\_rating": 4.0\} \\ 
    \hline
    \textbf{Claude-3.5-Sonnet} & \{"predicted\_rating": 3\} \\ 
    \hline
    \textbf{Ground Truth} & actual rating: 5.0 \\
     \bottomrule[1.5pt]
\end{tabular}
\end{adjustbox}
\label{tab:dolma_ugc1}
\end{table*}

\begin{table*}[t]
\caption{Pairwise ranking output in \ourmethod{}.}
\begin{adjustbox}{max width=1\linewidth}
\begin{tabular}{p{1.5in}p{5in}}
    \toprule[1.5pt]
    \textbf{Input} &  [User A]\newline
<|The Start of User Data|>\newline
\#\# User Profile\newline
The user appears to have a diverse range of interests and preferences based on their purchase history. They seem to be interested in kitchen and household items, with purchases including a foil cutter, cheese slicer, dish soap dispenser, trash can, and cabinet organizers. They also appear to enjoy beer and have purchased beer-related products like a beer dispenser and bar towel.\newline

The user's purchase history suggests an interest in home decor and organization, with items like decorative wall art, a vanity, and storage shelves. They have also purchased alarm clocks, indicating a need for timekeeping devices.\newline

In terms of electronics and appliances, the user has bought a Keurig coffee maker, a gaming chair, and a robot vacuum cleaner, suggesting an interest in convenience and technology.\newline

The user's ratings reveal a preference for high-quality and functional products, with items like the foil cutter, cheese slicer, vanity, and oscillating fan receiving high scores. However, they also seem to be dissatisfied with some purchases, as evidenced by the low scores given to certain items like alarm clocks and towels.\newline

Overall, the user appears to be practical and value-conscious, seeking products that serve specific purposes and offer good quality and functionality. Their interests span across various categories, including kitchen, home organization, decor, technology, and entertainment.\newline

\#\# User Most Common Rating\newline
1.0\newline

\#\# User Average Rating\newline
2.1\newline

\#\#\# Item Title\newline
Keurig K-Classic Coffee Maker K-Cup Pod, Single Serve, Programmable, 6 to 10 oz. Brew Sizes, Black\newline
\#\#\# User Rating\newline
1.0

\#\#\# Item Title\newline
Leick Favorite Finds Coffee Table\newline
\#\#\# User Rating\newline
1.0\newline

\#\#\# Item Title\newline
Ottomanson CTW1008-16X30 8 Piece Turkish Cotton Towels, 16" X 30"-Set of 6, Brown\newline
\#\#\# User Rating\newline
2.0\newline

\#\#\# Item Title\newline
OKP K8 Robot Vacuum and Mop Combo, 2000Pa Super Suction, Integrated Design of Dust Box Water Tank, Self Charging, Robotic Vacuums for Pet Hair, Blue\newline
\#\#\# User Rating\newline
1.0\newline
<|The End of User Data|>\\
     \bottomrule[1.5pt]
\end{tabular}
\end{adjustbox}
\label{tab:dolma_ugc1}
\end{table*}

\begin{table*}[t]
\caption{Pairwise ranking output in \ourmethod{}.}
\begin{adjustbox}{max width=1\linewidth}
\begin{tabular}{p{1.5in}p{5in}}
    \toprule[1.5pt]
    \textbf{Input} &  
[User B]\newline
<|The Start of User Data|>\newline
\#\# User Profile\newline
The user appears to have a strong interest in home decor and organization. They have purchased various items such as wall art, a headboard, a folding table, bed sheets, and storage ottomans, suggesting a desire to create a comfortable and stylish living environment. Additionally, their purchase of a vacuum cleaner and coat rack indicates a concern for cleanliness and organization.\newline

The user also seems to have a preference for quality and durability, as evidenced by their purchase of high-thread-count sheets and a stainless steel coffee percolator. They may value products that are long-lasting and well-made.\newline

Furthermore, the user has shown an interest in area rugs, with purchases ranging from small runners to larger rugs, potentially indicating a desire to add warmth and texture to their living spaces.\newline

Overall, the user profile suggests an individual who values a well-organized and aesthetically pleasing home environment, with a focus on quality and functional pieces that contribute to both comfort and style.\newline

\#\# User Most Common Rating\newline
5.0\newline

\#\# User Average Rating\newline
3.9\newline

\#\#\# Item Title\newline
West Bend 54159 Classic Stainless Steel Electric Coffee Percolator with Heat Resistant Handle and Base Features Detachable Cord, 12-cup, Silver\newline
\#\#\# User Rating\newline
2.0\newline

\#\#\# Item Title\newline
Cloth Napkins Set of 12 Cotton Linen Blend Printed Dinner Napkins Perfect for Parties Dinners Weddings Cocktail Christmas Napkins Cloth 20x20 Blue Floral\newline
\#\#\# User Rating\newline
5.0\newline

\#\#\# Item Title\newline
Weavric Cloth Dinner Napkin Bulk, Set of 12, 20 X 20 Inches Wrinkle-Free Washable Reusable Forest Green Linen Table Napkins with Hemmed Edge for Wedding, Party, Hotel, Restaurant\newline
\#\#\# User Rating\newline
5.0\newline

\#\#\# Item Title\newline
Crown Mark Barlow Bicast Headboard, King\newline
\#\#\# User Rating\newline
5.0\newline

<|The End of User Data|>\\

     \bottomrule[1.5pt]
\end{tabular}
\end{adjustbox}
\label{tab:dolma_ugc1}
\end{table*}

\begin{table*}[t]
\caption{Pairwise ranking output in \ourmethod{}.}
\begin{adjustbox}{max width=1\linewidth}
\begin{tabular}{p{1.5in}p{5in}}
    \toprule[1.5pt]
    \textbf{Input} &  [Query Item Details]\newline
<|The Start of Query Item Information|>\newline
\#\#\# Item Title\newline
Keurig K-Slim Coffee Maker, Single Serve K-Cup Pod Coffee Brewer, Multistream Technology, Scarlet Red\newline
<|The End of Query Item Information|>\newline

Which user prefer the query item more? Output only "[User A]" or "[User B]", do not generate anything else:\\
    \hline 
    \textbf{Gemma-2-9B-it} & [User A] \\
    \hline
    \textbf{Qwen-2.5-14B-it} & [User B]\\
    \hline 
    \textbf{Llama-3.1-70B-it}& [User A]\\
    \hline
    \textbf{GPT-4o-mini} & [User B] \\  
    \hline
    \textbf{Claude-3.5-Sonnet} & [User B] \\ 
    \hline
    \textbf{Ground Truth} & [User B]\\
     \bottomrule[1.5pt]
\end{tabular}
\end{adjustbox}
\label{tab:dolma_ugc1}
\end{table*}

\begin{table*}[t]
\caption{Listwise ranking output in \ourmethod{}.}
\begin{adjustbox}{max width=1\linewidth}
\begin{tabular}{p{1.5in}p{5in}}
    \toprule[1.5pt]
    \textbf{Input} & [Users]\newline<|The Start of User1 Data|>\newline\#\# User Profile\newline Based on the user behavior history data, this user appears to have a strong interest in kitchen and household items. They have purchased a variety of appliances and tools for cooking, baking, and food preparation, such as a stand mixer, sandwich maker, grill, and coffee maker. Additionally, they seem to value convenience and practicality, as evidenced by their purchase of a touchless trash can and a slim, shatterproof pitcher. The user also seems to appreciate furniture and decor items that have a rustic or natural aesthetic, as shown by their purchase of a rustic end table. Overall, this user likely enjoys cooking, entertaining, and creating a comfortable and functional living space.\newline\newline\#\# User Most Common Rating\newline5.0\newline\newline\#\# User Average Rating\newline4.5\newline\newline\#\#\# Item Title\newline Keurig K-Mini Plus Coffee Maker, Single Serve K-Cup Pod \& Keurig K-Cup Pod \& Ground Coffee Storage Unit\newline\#\#\# User1 Rating\newline5.0\newline\newline\#\#\# Item Title\newline Keurig K-Mini Plus Coffee Maker, Single Serve K-Cup Pod \& Keurig K-Cup Pod \& Ground Coffee Storage Unit\newline\#\#\# User1 Rating\newline5.0\newline\newline\#\#\# Item Title\newline Signature Design by Ashley - Mestler Rustic Chairside End Table w/ Two Fixed Multi-Colored Shelves, Brown\newline\#\#\# User1 Rating\newline5.0\newline\newline\#\#\# Item Title\newline Signature Design by Ashley - Mestler Rustic Chairside End Table w/ Two Fixed Multi-Colored Shelves, Brown\newline\#\#\# User1 Rating\newline5.0\newline<|The End of User1 Data|>\newline\newline<|The Start of User2 Data|>\newline\#\# User Profile\newline The user appears to have a strong interest in home decor and organization. They have purchased various items such as wall art, a headboard, a folding table, bed sheets, and storage ottomans, suggesting a desire to create a comfortable and stylish living environment. Additionally, their purchase of a vacuum cleaner and coat rack indicates a concern for cleanliness and organization.\newline\newline The user also seems to have a preference for quality and durability, as evidenced by their purchase of high-thread-count sheets and a stainless steel coffee percolator. They may value products that are long-lasting and well-made. \\
     \bottomrule[1.5pt]
\end{tabular}
\end{adjustbox}
\label{tab:dolma_ugc1}
\end{table*}

\begin{table*}[t]
\caption{Listwise ranking output in \ourmethod{}.}
\begin{adjustbox}{max width=1\linewidth}
\begin{tabular}{p{1.5in}p{5in}}
    \toprule[1.5pt]
    \textbf{Input} & Furthermore, the user has shown an interest in area rugs, with purchases ranging from small runners to larger rugs, potentially indicating a desire to add warmth and texture to their living spaces.\newline\newline Overall, the user profile suggests an individual who values a well-organized and aesthetically pleasing home environment, with a focus on quality and functional pieces that contribute to both comfort and style.\newline\newline\#\# User Most Common Rating\newline5.0\newline\newline\#\# User Average Rating\newline3.9\newline\newline\#\#\# Item Title\newline West Bend 54159 Classic Stainless Steel Electric Coffee Percolator with Heat Resistant Handle and Base Features Detachable Cord, 12-cup, Silver\newline\#\#\# User2 Rating\newline2.0\newline\newline\#\#\# Item Title\newline Cloth Napkins Set of 12 Cotton Linen Blend Printed Dinner Napkins Perfect for Parties Dinners Weddings Cocktail Christmas Napkins Cloth 20x20 Blue Floral\newline\#\#\# User2 Rating\newline5.0\newline\newline\#\#\# Item Title\newline Weavric Cloth Dinner Napkin Bulk, Set of 12, 20 X 20 Inches Wrinkle-Free Washable Reusable Forest Green Linen Table Napkins with Hemmed Edge for Wedding, Party, Hotel, Restaurant\newline\#\#\# User2 Rating\newline5.0\newline\newline\#\#\# Item Title\newline Crown Mark Barlow Bicast Headboard, King\newline\#\#\# User2 Rating\newline5.0\newline<|The End of User2 Data|>\newline\newline<|The Start of User3 Data|>\newline\#\# User Profile\newline The user appears to have diverse interests spanning personal care, kitchen gadgets, and home organization. They seem to value quality and convenience, as evidenced by their high ratings for items like the vanilla sticks, manual food chopper, herb mincer, and Oster convection toaster oven. However, they also express dissatisfaction with certain products, such as the humidifier, milk frother, can openers, and pepper mill, suggesting a discerning eye for functionality.\newline The user's interest in kitchen tools and appliances is evident, with a focus on efficient food preparation and storage solutions. The purchase of a high-quality food processor with a spiralizer attachment indicates a potential interest in healthy eating or culinary exploration.\newline Organization and storage seem to be important to the user, as demonstrated by their purchase of a shoe storage rack and refrigerator liners. Comfort is also a consideration, with the purchase of a heated mattress pad and a leg elevation pillow, although the latter received a low rating.\\
     \bottomrule[1.5pt]
\end{tabular}
\end{adjustbox}
\label{tab:dolma_ugc2}
\end{table*}

\begin{table*}[t]
\caption{Listwise ranking output in \ourmethod{}.}
\begin{adjustbox}{max width=1\linewidth}
\begin{tabular}{p{1.5in}p{5in}}
    \toprule[1.5pt]
    \textbf{Input} & Overall, the user appears to be a practical and discerning consumer who values quality, convenience, and organization in their personal and kitchen-related purchases, while also exploring new culinary avenues and prioritizing comfort at home.\newline\newline\#\# User Most Common Rating\newline1.0\newline\newline\#\# User Average Rating\newline2.7\newline\newline\#\#\# Item Title\newline Best Cool Mist Humidifier UltraSonic Steam Vaporizer - Whisper Quiet Technology, Moistair Electric with Warm LED Light\newline\#\#\# User3 Rating\newline1.0\newline\newline\#\#\# Item Title\newline Cusinart Small Pusher\newline\#\#\# User3 Rating\newline5.0\newline\newline\#\#\# Item Title\newline ZYLISS FastCut Herb Mincer\newline\#\#\# User3 Rating\newline5.0\newline\newline\#\#\# Item Title\newline FRESH STICKS - Golden Vanilla\newline\#\#\# User3 Rating\newline5.0\newline<|The End of User3 Data|>\newline\newline<|The Start of User4 Data|>\newline\#\# User Profile\newline The user appears to have a diverse range of interests and preferences based on their purchase history. They seem to be interested in kitchen and household items, with purchases including a foil cutter, cheese slicer, dish soap dispenser, trash can, and cabinet organizers. They also appear to enjoy beer and have purchased beer-related products like a beer dispenser and bar towel.\newline\newline The user's purchase history suggests an interest in home decor and organization, with items like decorative wall art, a vanity, and storage shelves. They have also purchased alarm clocks, indicating a need for timekeeping devices.\newline\newline In terms of electronics and appliances, the user has bought a Keurig coffee maker, a gaming chair, and a robot vacuum cleaner, suggesting an interest in convenience and technology. \\
   
     \bottomrule[1.5pt]
\end{tabular}
\end{adjustbox}
\label{tab:dolma_ugc2}
\end{table*}

\begin{table*}[t]
\caption{Listwise ranking output in \ourmethod{}.}
\begin{adjustbox}{max width=1\linewidth}
\begin{tabular}{p{1.5in}p{5in}}
    \toprule[1.5pt]
    \textbf{Input} & The user's ratings reveal a preference for high-quality and functional products, with items like the foil cutter, cheese slicer, vanity, and oscillating fan receiving high scores. However, they also seem to be dissatisfied with some purchases, as evidenced by the low scores given to certain items like alarm clocks and towels.\newline \newline Overall, the user appears to be practical and value-conscious, seeking products that serve specific purposes and offer good quality and functionality. Their interests span across various categories, including kitchen, home organization, decor, technology, and entertainment.\newline\newline\#\# User Most Common Rating\newline1.0\newline\#\# User Average Rating\newline2.1\newline\#\#\# Item Title\newline Keurig K-Classic Coffee Maker K-Cup Pod, Single Serve, Programmable, 6 to 10 oz. Brew Sizes, Black\newline\#\#\# User4 Rating\newline1.0\newline\#\#\# Item Title\newline Leick Favorite Finds Coffee Table\newline\#\#\# User4 Rating\newline1.0\newline\#\#\# Item Title\newline Ottomanson CTW1008-16X30 8 Piece Turkish Cotton Towels, 16\" X 30\"-Set of 6, Brown\newline\#\#\# User4 Rating\newline2.0\newline\#\#\# Item Title\newline OKP K8 Robot Vacuum and Mop Combo, 2000Pa Super Suction, Integrated Design of Dust Box Water Tank, Self Charging, Robotic Vacuums for Pet Hair, Blue\newline\#\#\# User4 Rating\newline1.0\newline<|The End of User4 Data|>\newline\newline[Query Item Details]\newline<|The Start of Query Item Information|>\newline\#\#\# Item Title\newline Keurig K-Slim Coffee Maker, Single Serve K-Cup Pod Coffee Brewer, Multistream Technology, Scarlet Red\newline<|The End of Query Item Information|>\newline\newline The answer ranks users from most preferred to least preferred for the query item and adhere to the following JSON format, do not include any additional information: \{\"predicted\_ranking\": <user\_ranking>\}\newline Answer: \\
    \hline 
    \textbf{Gemma-2-9B-it} & \{"predicted\_ranking": [1, 2, 4, 3]\} \\
    \hline
    \textbf{Qwen-2.5-14B-it} & \{"predicted\_ranking": [1, 2, 4, 3]\} \\
    \hline 
    \textbf{Llama-3.1-70B-it}& \{"predicted\_ranking": [1, 2, 4, 3]\}\\
    \hline
    \textbf{GPT-4o-mini} & \{"predicted\_ranking": [1, 2, 3, 4]\}\\  
    \hline
    \textbf{Claude-3.5-Sonnet} & \{"predicted\_ranking": [1, 2, 3, 4]\} \\ 
    \hline
    \textbf{Ground Truth} & [2, 1, 4, 3]\\ 
     \bottomrule[1.5pt]
\end{tabular}
\end{adjustbox}
\label{tab:dolma_ugc2}
\end{table*}

\end{small}

\end{document}